\documentclass[times,onecolumn,final]{elsarticle}

\global\bibsep=0pt
\usepackage{graphicx,psfrag,epsf}
\usepackage{algorithm}
\usepackage{algorithmicx}
\usepackage{algpseudocode}

\floatname{algorithm}{Algorithm}
\usepackage{url, amssymb, amsfonts}
\usepackage{amsmath}
\usepackage{amsthm}
\theoremstyle{plain}

\newtheorem{corollary}{Corollary}

\usepackage{thmtools, thm-restate}

\usepackage{thmtools, thm-restate}

\providecommand{\sct}[1]{{\texttt{#1}}}
\newcommand{\Anova}{\sct{Anova}}
\newcommand{\Manova}{\sct{Manova}}
\newcommand{\Disco}{\sct{Disco}}
\newcommand{\Energy}{\sct{Energy}}
\newcommand{\Mmd}{\sct{Mmd}}
\newcommand{\Kmerf}{\sct{Kmerf}}
\newcommand{\Mgc}{\sct{Mgc}}

\newcommand{\Dcor}{\sct{Dcor}}
\newcommand{\Dcov}{\sct{Dcov}}
\newcommand{\Hsic}{\sct{Hsic}}

\newcommand{\RV}{\sct{Rv}}
\newcommand{\CCA}{\sct{Cca}}

\begin{document}
        
\begin{frontmatter}
\title{Universally Consistent K-Sample Tests via Dependence Measures}
\author[1]{Sambit Panda\corref{equal}}
\author[2]{Cencheng Shen\corref{equal}}
\author[1]{Ronan Perry}
\author[3]{Jelle Zorn}
\author[3]{Antoine Lutz}
\author[4]{Carey E.~Priebe}
\author[1,5]{Joshua T. Vogelstein\corref{cor1}}

\address[1]{Department of Biomedical Engineering, Johns Hopkins University, Maryland, USA}
\address[2]{Department of Applied Economics and Statistics, University of Delaware, Delaware, USA}
\address[3]{Lyon Neuroscience Research Centre, Universite Claude Bernard Lyon 1, Lyon, France}
\address[4]{Department of Applied Mathematics and Statistics, Johns Hopkins University, Maryland, USA}
\address[5]{Center for Imaging Science and Kavli~Neuroscience~Discovery Institute, Johns Hopkins University, Maryland, USA}

\cortext[equal]{Sambit Panda and Cencheng Shen contribute equally to this work.}
\cortext[cor1]{Corresponding author. Email: jovo@jhu.edu}

\begin{abstract}
The K-sample testing problem involves determining whether K groups of data points are each drawn from the same distribution. Analysis of variance is arguably the most classical method to test mean differences, along with several recent methods to test distributional differences. In this paper, we demonstrate the existence of a transformation that allows K-sample testing to be carried out using any dependence measure. Consequently, universally consistent K-sample testing can be achieved using a universally consistent dependence measure, such as distance correlation and the Hilbert-Schmidt independence criterion. This enables a wide range of dependence measures to be easily applied to K-sample testing. 
\end{abstract}
\end{frontmatter}
\section{Introduction}
\label{sec_intro}
Given two datasets $\{u_{i}^{(1)} \in \mathbb{R}^p, i=1,\ldots, n_1\}$ and $\{u_{j}^{(2)} \in \mathbb{R}^p, j=1,\ldots, n_2\}$, assume each $u_{i}^{(1)}$ is sampled independently and identically (i.i.d.)~from $F_{U_1}$, and each $u_{j}^{(2)}$ is sampled i.i.d.~from $F_{U_2}$. Also, assume that each pair $(u_{i}^{(1)},u_{j}^{(1)})$ is independent for any $(i,j)$. The classical two-sample testing problem tests whether the two datasets were sampled from the same distribution, stated as:
\begin{align*} 
H_0 &: F_{U_1} = F_{U_2}, \\
H_A &: F_{U_1} \neq F_{U_2}.
\end{align*}
The K-sample testing problem is a generalization of the above. Let $\{u_{i}^{(k)} \stackrel{i.i.d.}{\sim} F_{U_k} \in \mathbb{R}^p, i=1,\ldots,n_k\}$ for $k=1,\dots, K$, and assume $(u_{i}^{(s)},u_{j}^{(t)})$ is independent for any $(i,j,s,t)$. We aim to test:
\begin{align*} 
    H_0 &: F_{U_1} = F_2 = \cdots = F_{U_K}, \\
    H_A &: \exists \ s \neq t \text{ s.t. } F_{U_s} \neq F_{U_t}.
\end{align*}
Student's t-test and its multivariate generalization, Hotelling's $T^2$, are traditionally used for two-sample testing, while analysis of variance (\Anova) or multivariate analysis of variance (\Manova) are conventional choices for K-sample tests. These tests, however, only aim to test mean differences, and do not perform well for non-Gaussian data beyond their parametric assumptions \citep{warne2014primer}. To address this, several nonparametric statistics have been developed to test distributional differences, such as \Energy~\citep{szekely2013energy} and maximal mean discrepancy (\Mmd) \citep{gretton2012kernel} for two-sample tests, and multivariate Heller-Heller-Gorfine \citep{heller2016consistent} and distance components (\Disco) \citep{rizzo2010disco} for K-sample tests.

A closely related and popular problem in statistics is testing independence. Given $x_i \in \mathbb{R}^p$ and $y_i \in \mathbb{R}^q$, and $n$ samples of $(x_i, y_i) \overset{iid}{\sim} F_{XY}$, the independence hypothesis is stated as:
\begin{align*} 
    H_0 &: F_{XY} = F_X F_Y, \\
    H_A &: F_{XY} \neq F_X F_Y.
\end{align*}
Traditional Pearson's correlation \citep{pearson1895vii} is popular but limited to detecting linear dependence. Many non-parametric methods have been proposed recently, such as distance covariance (\Dcov) \citep{szekely2007measuring, szekely2013distance}, Hilbert-Schmidt independence criterion (\Hsic) \citep{gretton2005kernel, gretton2010consistent,Bounliphone2016}, multiscale graph correlation (\Mgc) \citep{MGC, MGCDCor}, among many others \citep{heller2012consistent, Pan2019, Zhou2024, DCorRF}.

These recent dependence measures are universally consistent under mild assumptions, such as finite moments. Specifically, the sample statistic converges to a population statistic, which equals $0$ if and only if $X$ and $Y$ are independent. Therefore, these measures can detect any type of relationship given a sufficiently large sample size, and the testing power increases to $1$ as $n$ increases, regardless of whether the underlying relationship is linear or nonlinear. These universally consistent dependence measures have found applications for various inference tasks, such as feature screening \citep{Li2012, Zhong2015, DCorScreening,DCorHD}, time-series analysis \citep{Zhou2012, DCorTemporal}, conditional independence \citep{Gretton2007, szekely2014partial, Wang2015}, and graph testing \citep{MGCGraph, DCorGraph}.

In this manuscript, we establish a fundamental connection between K-sample testing and independence testing: there exists a transformation of the given data such that the K-sample testing problem is converted to the independence testing problem on the transformed data. As a result, any universally consistent dependence measure can be used to achieve universally consistent K-sample testing. Moreover, the proposed sample transformation allows previously established two-sample statistics, such as \Energy~and \Mmd, to be equivalent to the corresponding dependence measures, \Dcov~and \Hsic, respectively; and the more general K-sample \Disco~equals a bootstrap version of \Dcov. Finally, we used simulated data to verify the validity, consistency, and finite-sample testing power of several popular dependence measures for K-sample testing. Theorem proofs and additional simulations are provided in the supplementary material. The code is available in the Hyppo statistical package \citep{DCorHyppo} and on GitHub\footnote{\url{https://hyppo.neurodata.io/}}.

\section{Method and Theory}
\label{sec_results}

In this section, we first review \Dcov~and \Hsic, which are the foundational blocks of universally consistent dependence measures. Next, we introduce the population transformation that converts K-sample testing to independence testing in the random variable setting, enabling any consistent dependence measure to achieve consistent K-sample testing. We then proceed to the sample method and prove the sample equivalence between \Dcov~and \Energy, and between \Hsic~and \Mmd. Throughout this section, we assume all distributions have finite moments, and all proofs are provided in the appendix.

\subsection{Review of Dependence Measures}
Denote the paired sample data as $(\mathbf{X},\mathbf{Y}) = \{(x_{i},y_{i}) \in \mathbb{R}^{p+q}, i=1,\ldots,n\}$, where each sample pair $(x_{i},y_{i})$ is assumed to be i.i.d.~as $F_{XY}$ with finite moments. Given a distance metric $d(\cdot,\cdot): \mathbb{R}^{p} \times \mathbb{R}^{q} \rightarrow \mathbb{R}$, such as the Euclidean metric, denote $\mathbf{D}^{\mathbf{X}}$ and $\mathbf{D}^{\mathbf{Y}}$ as the $n \times n$ distance matrices for $\mathbf{X}$ and $\mathbf{Y}$, respectively. Define $\mathbf{H} = \vec{I} - \frac{1}{n} \vec{J}$ as an $n \times n$ centering matrix, where $\vec{I}$ is the identity matrix and $\vec{J}$ is the matrix of ones. Then the sample distance covariance (\Dcov) and distance correlation (\Dcor) can be computed by:
\begin{align*}
\text{\Dcov}_{n}(\mathbf{X}, \mathbf{Y})&= \frac{1}{n^2}trace(\mathbf{H}\mathbf{D}^{\mathbf{X}}\mathbf{H}\mathbf{H}\mathbf{D}^{\mathbf{Y}}\mathbf{H}),\\
\text{\Dcor}_{n}(\mathbf{X}, \mathbf{Y})&= \frac{\text{\Dcov}_{n}(\mathbf{X}, \mathbf{Y})}{\sqrt{\text{\Dcov}_{n}(\mathbf{X}, \mathbf{X}) \text{\Dcov}_{n}(\mathbf{Y}, \mathbf{Y})}}.
\end{align*}
By default, distance correlation utilizes the Euclidean distance as its metric, but it can be any metric of strong negative type \citep{lyons2013distance}, or a characteristic kernel upon a kernel to metric transformation \citep{sejdinovic2013equivalence,DCorKernel}. Moreover, by replacing the distance metric $d(\cdot,\cdot)$ with a kernel measure $k(\cdot,\cdot)$, such that $\mathbf{D}^{\mathbf{X}}$ and $\mathbf{D}^{\mathbf{Y}}$ become the corresponding kernel matrices, \Dcov~becomes \Hsic. 

\subsection{Population Transformation}
\begin{restatable}{theorem}{thmOne}
\label{thm1}
Given $K$ random variables $(U_1, U_2, \ldots,U_K)$. Let $V \in \mathbb{R}^{K}$ be the multinomial distribution of probability $(\pi_1, \pi_2, \ldots,\pi_K)$, where $\pi_k \in (0,1)$ and $\sum_{k=1}^{K}\pi_k=1$. Let $U$ be the following mixture distribution:
\[
U = \sum_{k=1}^{K} 1(V_k=1) U_{k},
\]
where $V_k$ denotes the $k$th dimension of $V$. Then, $F_{UV} = F_U F_V$ if and only if $F_{U_1} = F_{U_2} = \cdots = F_{U_K}$.
\end{restatable}
Therefore, the proposed transformation from $(U_1, U_2, \ldots, U_K) \rightarrow (U, V)$ converts K-sample testing on $(U_1, U_2, \ldots, U_K)$ to independence testing on $(U, V)$, leading to the consistency of dependence measures for K-sample testing.

\begin{corollary}
\label{cor1} 
Suppose $\tau(\cdot, \cdot)$ is a universally consistent dependence measure such that $\tau(X, Y) = 0$ if and only if $X$ and $Y$ are independent. Then, via the proposed transformation in Theorem~\ref{thm1}, $\tau(U, V)$ is universally consistent for the K-sample test on ${U_k}$, i.e., $\tau(U, V) = 0$ if and only if $F_{U_1} = F_{U_2} = \cdots = F_{U_K}$.
\end{corollary}

Note that the probabilities $\{\pi_k\}$ can be easily chosen based on sample size, as described in the sample method subsection below. However, the transformation and the resulting consistency apply to any $\{\pi_k\}$, as long as all probabilities are positive, ensuring that each random variable $U_k$ has a positive probability of appearing in the mixture $U$. For example, if we use \Dcov~as the statistic, the value of \Dcov$(U, V)$ may differ for different choices of $\{\pi_k\}$ when some distributions are different. Nonetheless, \Dcov$(U, V) = 0$ if and only if all distributions are the same, which holds for any choice of $\{\pi_k\}$.

\subsection{Sample Method}

Given the sample data and a sample dependence measure $\tau_{n}(\cdot,\cdot)$, we can carry out the K-sample testing as follows:

\begin{itemize}
\item \textbf{Input}: For each $k=1,\ldots,K$, the sample data $\mathbf{U}_k=[u_1^{(k)},\ldots,u_{n_k}^{(k)}]^{T} \in \mathbb{R}^{n_k \times p}$; a given sample dependence measure $\tau_{n}(\cdot,\cdot): \mathbb{R}^{n \times p} \times \mathbb{R}^{n \times K} \rightarrow \mathbb{R}$.
\item \textbf{Sample Transformation}: Let $n = \sum_{k=1}^K n_k$, we set $\mathbf{U} \in \mathbb{R}^{n \times p}$ as the row-concatenation of all data matrices, and $\mathbf{V} \in \mathbb{R}^{n \times K}$ as a one-hot encoding of the data source label. That is,
\begin{align*}
\mathbf{U} &= \begin{bmatrix} \mathbf{U}_1 \\ \vdots \\ \mathbf{U}_K \end{bmatrix} \in \mathbb{R}^{n \times p}, \\
\mathbf{V} &= 
\begin{bmatrix}
    \vec{1}_{n_1 \times 1} & \vec{0}_{n_1 \times 1} & \ldots & \vec{0}_{n_1 \times 1} \\
    \vec{0}_{n_2 \times 1} & \vec{1}_{n_2 \times 1} & \ldots & \vec{0}_{n_2 \times 1} \\
    \vdots & \vdots & \ddots & \vdots \\
    \vec{0}_{n_K \times 1} & \vec{0}_{n_K \times 1} & \ldots & \vec{1}_{n_K \times 1} \\
\end{bmatrix} \in \mathbb{R}^{n \times K},
\end{align*}
where $\vec{1}_{n_1 \times 1}$ is the vector of ones of length $n_1$. 
\item \textbf{Dependence Measure}: Compute $\tau_{n}(\mathbf{U},\mathbf{V})$.
\item \textbf{Permutation Test}: Compute the p-value via a standard permutation test, i.e., 
\begin{align*}
\mbox{pval}= \frac{1}{R}\sum_{r=1}^{R} 1\{\tau_n(\mathbf{U},\mbox{perm}_r(\mathbf{V}))> \tau_n(\mathbf{U},\mathbf{V})\}
\end{align*}
where $\mbox{perm}_r()$ represents a random permutation of size $n$ (which is different and randomized for each $r$), and $R$ is the number of permutations.
\item \textbf{Output}: The sample statistic $\tau_{n}(\mathbf{U},\mathbf{V})$ and its p-value. 
\end{itemize}

Here, the sample matrix pair $(\mathbf{U}, \mathbf{V})$ can be viewed as the sample realization of the population transformation $(U, V)$, where the mixture probability $\{\pi_k = \frac{n_k}{n}, k=1,\ldots,K\}$. The one-hot encoding scheme has been a fundamental technique in neural networks and machine learning \citep{Bishop1995,Murphy2012} and has recently been applied in graph embedding \citep{GEEOne,GEETemporalDynamics,GEEFusionGraphs}. 
For the choice of dependence measure, $\tau_{n}(\mathbf{U}, \mathbf{V})$ can be any aforementioned sample dependence measure, such as \Dcov, \Hsic, \Mgc, etc. While the permutation test is used here, one could use a faster testing procedure, such as the chi-square test via an unbiased test statistic \citep{DCorFast}. 

Because $\mathbf{V}$ is categorical, the distance between any two rows of $\mathbf{V}$ can only take two values. Specifically, $d(\mathbf{V}(i,:), \mathbf{V}(j,:))$ is either $0$ when $\mathbf{V}(i,:) = \mathbf{V}(j,:)$ or $\sqrt{2}$ when $\mathbf{V}(i,:) \neq \mathbf{V}(j,:)$ under Euclidean distance. The former occurs when the $i$th and $j$th sample data come from the same group, and the latter occurs when they come from different groups. We use $\beta$ to denote the maximum distance minus the minimum distance within the distance matrix of $\mathbf{V}$. As the first and last observations in $\mathbf{V}$ always come from different groups based on our construction, we can conveniently let $\beta = d(\mathbf{V}(1,:), \mathbf{V}(n,:))$ in this case, where $\mathbf{V}(1, :)$ represents the first row and $\mathbf{V}(n, :)$ represents the last row of the matrix.

\subsection{Sample Properties}
Given the above sample transformation, the sample distance covariance for $(\mathbf{U}, \mathbf{V})$ can be proved to be exactly the same as the sample energy statistic for $(\mathbf{U}_1, \mathbf{U}_2)$, up to a scalar constant.

\begin{restatable}{theorem}{thmTwo}
\label{thm2}
Assume a translation-invariant metric $d(\cdot,\cdot)$ is used, and denote $\beta=d(\mathbf{V}(1,:),\mathbf{V}(n,:))$. It follows that
\[
\text{\Dcov}_n (\mathbf{U}, \mathbf{V}) =\frac{2n_1^2 n_2^2 \beta}{n^4} \cdot \text{\Energy}_{n_1,n_2} (\mathbf{U}_1, \mathbf{U}_2), 
\]
Under the permutation test, distance covariance, distance correlation, and energy statistic have the same testing p-value.
\end{restatable}

By default, both \Dcov~and \Energy~use Euclidean distance, which is translation-invariant. Moreover, due to the existing transformation between distance and kernel, \Hsic~is also equivalent to \Mmd.
\begin{restatable}{theorem}{thmThree}
\label{thm3}
Assume a translation-invariant kernel $k(\cdot,\cdot)$ is used, and 
denote $\beta=k(\mathbf{V}(1,:),\mathbf{V}(n,:))-k(\mathbf{V}(1,:),\mathbf{V}(1,:))$. It follows that
\[
\text{\Hsic}_n (\mathbf{U}, \mathbf{V}) =\frac{2n_1^2 n_2^2 \beta}{n^4} \cdot \text{\Mmd}_{n_1,n_2} (\mathbf{U}_1, \mathbf{U}_2), 
\]
Under the permutation test, Hilbert-Schmidt independence criterion and maximum mean discrepancy have the same testing p-value.
\end{restatable}

Lastly, in the case of general K-sample testing, the sample distance covariance for $(\mathbf{U},\mathbf{V})$ is a weighted summation of pairwise two-sample energy statistics. This coincides with the K-sample \Disco~statistic if the data sources are equally weighted.

\begin{restatable}{theorem}{thmFour}
\label{thm4}
Assume a translation-invariant metric $d(\cdot,\cdot)$ is used, and denote $\beta=d(\mathbf{V}(1,:),\mathbf{V}(n,:))$. It follows that
\[
\begin{aligned}
\text{\Dcov}_{n}(\mathbf{U}, \mathbf{V}) = \beta\sum_{1 \leq s<t \leq K}&\left\{\frac{n (n_s+n_t)-\sum_{l=1}^{K} n_{l}^2}{n^4}\cdot n_s n_t  \text{\Energy}_{n_s,n_t}(\mathbf{U}_s, \mathbf{U}_t) \right\}.
\end{aligned}
\]
Moreover, sample distance covariance is equivalent to the sample \Disco~statistic when $n_1 = n_2 =\ldots=n_k$, in which case 
\[
\text{\Dcov}_{n}(\mathbf{U}, \mathbf{V}) =\frac{2 \beta}{n K} \text{\Disco}(\{\mathbf{U}_k\}).
\]
\end{restatable}
Therefore, if we consider a bootstrap resampling of the given data $\{\mathbf{U}_k\}$, such that the bootstrap samples consist of the same number of observations per data source, and let $\mathbf{U}$ and $\mathbf{V}$ be the concatenated bootstrap samples, this will enforce $n_1 = n_2 = \ldots = n_K$. Thus, distance covariance for K-sample testing is equivalent to \Disco~on equally-weighted bootstrap samples.

It is worth noting that Proposition 6 and Corollary 5 of \cite{Edelmann2022} also establish the equivalence between \Hsic~and \Mmd, which aligns with our Theorem~\ref{thm3} for a kernel choice where $\beta=1$. Their work focused on discrete kernels for categorical data and found that \Hsic~coincides with \Mmd~in this context. In contrast, our paper employs one-hot encoding to transform the K-sample problem into an independence test, with $\mathbf{V}$ always being discrete. Interestingly, despite starting from different perspectives and following different procedures, we arrived at the same equivalence.

\section{Simulations}

We aim to verify the validity and consistency of using dependence measures for two-sample tests. Specifically, we consider three univariate settings and compare \Anova, distance correlation, and \Hsic~in each case. Note that more numerical comparisons involving other dependence measures and non-Gaussian simulations are provided in appendix Figure~\ref{fig1}.

We set up two random variables $U_1$ and $U_2$ and the corresponding parameters as follows:
\begin{enumerate}
\item \textbf{Sample Size Difference}: Both $U_1$ and $U_2$ are standard normal. $n_1=100$, and $n_2=20,40,\ldots,200$.
\item \textbf{Mean Difference}: $U_1$ is standard normal, and $U_2 \sim \mbox{Normal}(c, 1)$, where $c=0,0.05,0.1,\ldots,0.5$. $n_1=100$, and $n_2=200$.
\item \textbf{Variance Difference}: $U_1$ is standard normal, and $U_2 \sim \mbox{Normal}(0, 1+c)$, where $c=0,0.1,\ldots,1$. $n_1=100$, and $n_2=200$.
\end{enumerate}
We generate sample data $\mathbf{U}_1$ and $\mathbf{U}_2$ accordingly, and compute the p-value for each method. This is repeated for $1000$ replicates, and we compute the testing power of each method by checking how often the p-value is lower than at the type 1 error level $0.05$.

\begin{figure}
    \begin{center}
    \includegraphics[width=0.99\linewidth]{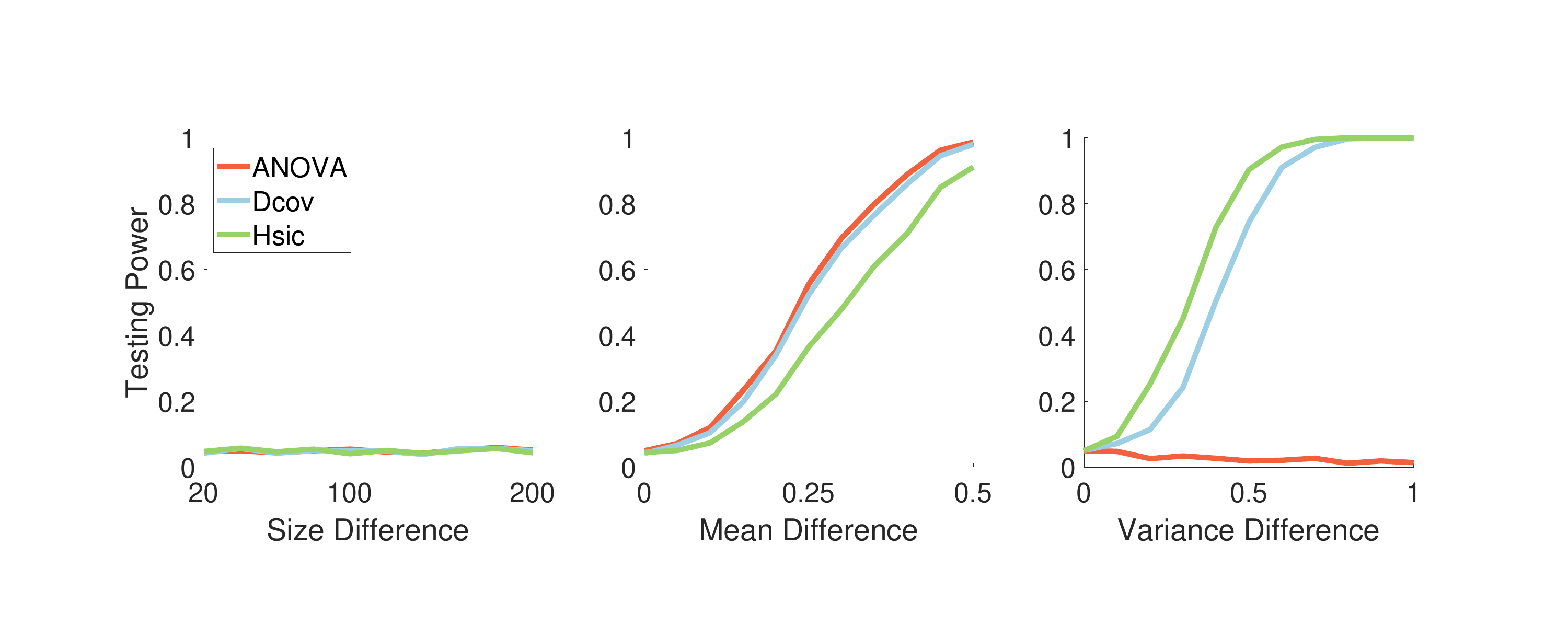}
    \end{center}
    \caption{The figure compares the testing power of \Anova, \Dcov, and \Hsic~for three different Gaussian-simulated sample datasets. }
    \label{fig4}
\end{figure}

Figure \ref{fig4} shows the testing power for each simulation. In the first simulation (left panel), there is no distribution difference; only the sample sizes are different, and all methods have a testing power of about $0.05$. In the second simulation (center panel), the mean difference is detected by all three methods, with \Anova~and \Dcov~showing the best power, followed by \Hsic. In the last simulation (right panel), \Anova~has little power, while \Hsic~performs the best, followed by \Dcov.

Overall, the first simulation demonstrates the validity of all methods, while the second and third simulations demonstrate the consistency of dependence measures in testing distributional differences. As \Anova~is designed to detect mean differences in Gaussian settings, it works for the second simulation but not the third. We also notice that \Dcov~is better at detecting mean differences, while \Hsic~is better at detecting variance differences, which can be attributed to the fact that \Dcov~has better finite-sample power for detecting linear dependence, while \Hsic~is better at detecting nonlinear dependence. We note that while universally consistent dependence measures are guaranteed to achieve perfect testing power given a sufficiently large sample size, different dependence measures can excel at detecting different distributions, requiring less sample size to achieve power 1.

\section*{Acknowledgments}
This work is supported by the Defense Advanced Research Projects Agency (DARPA) Lifelong Learning Machines program through contract FA8650-18-2-7834, the National Institutes of Health awards RO1MH120482 and T32GM119998, and the National Science Foundation award DMS-1921310 and DMS-2113099. The authors would like to thank the editor, two anonymous reviewers, Dr. Russell Lyons, Dr. Minh Tang, Mr. Ronak Mehta, Mr. Eric W. Bridgeford, and the rest of the NeuroData group at Johns Hopkins University for valuable feedback.



\bibliographystyle{elsarticle-harv}
\bibliography{references}

\clearpage
\appendix
\setcounter{figure}{0}
\renewcommand{\thefigure}{F\arabic{figure}}
\renewcommand{\theequation}{E\arabic{equation}}
\renewcommand{\thesubsection}{\thesection.\arabic{subsection}}
\renewcommand{\thesubsubsection}{\thesubsection.\arabic{subsubsection}}
\pagenumbering{arabic}
\renewcommand{\thepage}{\arabic{page}}

\bigskip
\begin{center}
{\large\bf APPENDIX}
\end{center}

\section{Theorem Proofs}
\label{sec:proofs}

\thmOne*
\begin{proof}
When $F_{U_1} = F_{U_2} = \cdots = F_{U_K}$, it is immediate that the mixture distribution satisfies:
\[
U = \sum_{k=1}^{K} 1(V_k=1) U_{k} = U_1. 
\]
Namely, regardless of $V$, $U$ always has the same distribution. Therefore, $U$ and $V$ are independent and $F_{UV}=F_{U}F_{V}$.

Next, suppose $U$ and $V$ are independent, and $F_{UV}=F_{U}F_{V}$. Then $F_{U|V_k=1}=F_{U}$ for any $k$, such that 
\[
U_1 = U_2 = \cdots = U_K = U. 
\]
Therefore, $F_{U_1} = F_{U_2} = \cdots = F_{U_K}$.

For Corollary~\ref{cor1}: by Theorem~\ref{thm1}, when $F_{U_1} = F_{U_2} = \cdots = F_{U_K}$, we have $F_{UV} = F_{U} F_{V}$ and thus $\tau(U, V) = 0$. On the other hand, when $\tau(U, V) = 0$, it holds that $F_{UV} = F_{U} F_{V}$, leading to $F_{U_1} = F_{U_2} = \cdots = F_{U_K}$.
\end{proof}

\thmTwo*
\begin{proof}
Given any two sample data matrices $\mathbf{U}_1 \in \mathbb{R}^{n_1 \times p}$ and $\mathbf{U}_2 \in \mathbb{R}^{n_2 \times p}$, the two-sample energy statistic is defined as
\[
\text{\Energy}_{n_1,n_2}( \mathbf{U}_{1}, \mathbf{U}_{2})=\frac{1}{n_1^{2}n_2^{2}}  \left(2n_1n_2\sum_{i=1}^{n_1}\sum_{j=1}^{n_2}d(u_{i}^{(1)},u_{j}^{(2)})-n_2^2\sum_{i,j=1}^{n_1}d(u_{i}^{(1)},u_{j}^{(1)})-n_1^2\sum_{i,j=1}^{n_2}d(u_{i}^{(2)},u_{j}^{(2)})\right).
\]

Using the proposed sample transformation to yield $\mathbf{U} \in \mathbb{R}^{n \times p}$ and $\mathbf{V} \in \mathbb{R}^{n \times K}$, and let $\mathbf{D}^{\mathbf{U}}$ and $\mathbf{D}^{\mathbf{V}}$ be their $n \times n$ distance matrices, respectively. Recall $\mathbf{H} = \vec{I} - \frac{1}{n} \vec{J}$ is the $n \times n$ centering matrix, and the sample distance covariance follows as:
\begin{align*}
\text{\Dcov}_{n}(\mathbf{U}, \mathbf{V})&= \frac{1}{n^2}trace(\mathbf{H}\mathbf{D}^{\mathbf{U}}\mathbf{H}\mathbf{H}\mathbf{D}^{\mathbf{V}}\mathbf{H}) \\
&= \frac{1}{n^2}trace(\mathbf{D}^{\mathbf{U}}\mathbf{H}\mathbf{D}^{\mathbf{V}}\mathbf{H}) \\
&=\frac{1}{n^2}\sum\limits_{i,j=1}^{n} \mathbf{D}^{\mathbf{U}}_{ij} \cdot (\mathbf{H}\mathbf{D}^{\mathbf{V}}\mathbf{H})_{ij} 
\end{align*}
by the property of matrix trace and the idempotent property of the centering matrix $\mathbf{H}$. The two distance matrices satisfy
\begin{align*}
&\mathbf{D}^{\mathbf{U}}_{ij}=d(\mathbf{U}(i,:),\mathbf{U}(j,:)) = \begin{cases}
d(u_{i}^{(1)},u_{j}^{(1)})       & \text{if } 1 \leq i,j\leq n_1, \\
d(u_{i}^{(2)},u_{j}^{(2)})       & \text{if } n_1 < i,j\leq n, \\
d(u_{i}^{(1)},u_{j}^{(2)})       & \text{otherwise,}
\end{cases}
\\
&\mathbf{D}^{\mathbf{V}}_{ij}=d(\mathbf{V}(i,:),\mathbf{V}(j,:)) = \begin{cases}
0       & \text{if } 1 \leq i,j\leq n_1 \mbox{ or } n_1< i,j \leq n,\\
\beta  & \text{otherwise.}
\end{cases}
\end{align*} 
It follows that
\[
(\mathbf{H}\mathbf{D}^{\mathbf{V}}\mathbf{H})_{ij}=\begin{cases}
\frac{-2n_2^{2}}{n^2}\beta       & \text{if } 1 \leq i,j\leq n_1 ,\\
\frac{-2n_1^{2}}{n^2}\beta   & \text{if } n_1< i,j \leq n,\\
\frac{2n_1n_2}{n^2}\beta  & \text{otherwise}.
\end{cases}
\]
Therefore, up to a scaling factor, the centering scheme via distance covariance matches the weight of the energy statistic for each term. Expanding all terms leads to
\begin{align*}
\Dcov_{n}(\mathbf{U}, \mathbf{V})&=
\frac{\beta}{n^4}\left(4n_1n_2\sum_{i=1}^{n_1}\sum_{j=1}^{n_2}d(u_{i}^{(1)},u_{j}^{(2)})-2n_2^2\sum_{i,j=1}^{n_1}d(u_{i}^{(1)},u_{j}^{(1)})-2n_1^2\sum_{i,j=1}^{n_2}d(u_{i}^{(2)},u_{j}^{(2)})\right) \\
&=\frac{2n_1^2 n_2^2\beta}{n^4} \Energy_{n_1,n_2}( \mathbf{U}_{1}, \mathbf{U}_{2}).
\end{align*}

As the scalar $\frac{2n_1^2 n_2^2\beta}{n^4}$ is invariant under any permutation of the given sample data, distance covariance and energy statistic have the same testing p-value via a permutation test. Note that the above result holds for any translation-invariant metric $d(\cdot,\cdot)$. 
\end{proof}

\thmThree*
\begin{proof}
The equivalence between Hilbert-Schmidt independence criterion and maximum mean discrepancy can be established via the exact same proof as above, through the distance to kernel equivalence in \cite{DCorKernel}: Assuming $d(\cdot,\cdot)$ is replaced by a translation-invariant kernel $k(\cdot,\cdot)$, and the distance matrices $\mathbf{D}^{\mathbf{U}}$ and $\mathbf{D}^{\mathbf{V}}$ are now kernel matrices, we have $\text{\Energy}_{n_1,n_2}( \mathbf{U}_{1}, \mathbf{U}_{2})$ becomes $-\Mmd_{n_1,n_2}( \mathbf{U}_{1}, \mathbf{U}_{2})$ and $\Dcov_{n}( \mathbf{U}, \mathbf{V})$ becomes $-\Hsic_{n}( \mathbf{U}, \mathbf{V})$.

Every other step in the proof of Theorem~\ref{thm2} carries over. There is only a minor notation difference, because when $d(\cdot,\cdot)$ becomes a kernel $k(\cdot,\cdot)$, we have
\begin{align*}
\mathbf{D}^{\mathbf{V}}_{ij}=k(\mathbf{V}(i,:),\mathbf{V}(j,:)) = \begin{cases}
k(\mathbf{V}(1,:),\mathbf{V}(1,:))       & \text{if } 1 \leq i,j\leq n_1 \mbox{ or } n_1< i,j \leq n,\\
k(\mathbf{V}(1,:),\mathbf{V}(n,:))  & \text{otherwise.}
\end{cases}
\end{align*} 
Then
\[
(\mathbf{H}\mathbf{D}^{\mathbf{V}}\mathbf{H})_{ij}=\begin{cases}
\frac{-2n_2^{2}}{n^2}\beta       & \text{if } 1 \leq i,j\leq n_1 ,\\
\frac{-2n_1^{2}}{n^2}\beta   & \text{if } n_1< i,j \leq n,\\
\frac{2n_1n_2}{n^2}\beta  & \text{otherwise},
\end{cases}
\]
where $\beta=k(\mathbf{V}(1,:),\mathbf{V}(n,:)) - k(\mathbf{V}(1,:),\mathbf{V}(1,:))$. This is because $d(\mathbf{V}(1,:),\mathbf{V}(1,:))$ is always $0$ for a valid translation-invariant metric, in which case it was reduced to $\beta=d(\mathbf{V}(1,:),\mathbf{V}(n,:))$ in the proof of Theorem~\ref{thm2}. But in case of kernels, the kernel $k(\mathbf{V}(1,:),\mathbf{V}(1,:))$ is not necessarily $0$.
\end{proof}

\thmFour*
\begin{proof}
The K-sample \Disco~equals:
\[
\Disco(\{\mathbf{U}_{k}\}) = \sum\limits_{1 \leq s<t \leq K}\left\{ \frac{n_s n_t}{2n}\Energy_{n_s,n_t}(\mathbf{U}_{s}, \mathbf{U}_{l}) \right\},
\]
where each pairwise energy statistic equals
\[
\text{\Energy}_{n_s,n_t}( \mathbf{U}_{s}, \mathbf{U}_{t})=\frac{2}{n_s n_t}  \sum_{i=1}^{n_a}\sum_{j=1}^{n_t}d(u_{i}^{(s)},u_{j}^{(t)})-\frac{1}{n_s^2}\sum_{i,j=1}^{n_s}d(u_{i}^{(s)},u_{j}^{(s)})-\frac{1}{n_t^2}\sum_{i,j=1}^{n_t}d(u_{i}^{(t)},u_{j}^{(t)}).
\]

From the proof of Theorem~\ref{thm2}, the sample distance covariance equals
\begin{align*}
\text{\Dcov}_{n}(\mathbf{U}, \mathbf{V})=\frac{1}{n^2}\sum\limits_{i,j=1}^{n} \mathbf{D}^{\mathbf{U}}_{ij} \cdot (\mathbf{H}\mathbf{D}^{\mathbf{V}}\mathbf{H})_{ij}.
\end{align*}
Assuming observation $i$ comes from $s$th data and observation $j$ comes from $t$th data source, we have $\mathbf{D}^{\mathbf{U}}_{ij}=d(u_{i}^{(s)},u_{j}^{(t)})$, and $\mathbf{D}^{\mathbf{V}}$ equals
\[
\mathbf{D}^{\mathbf{V}}_{ij} = \begin{cases}
0       & \text{if $s=t$},\\
\beta  & \text{if $s \neq t$}.
\end{cases}
\]
We then check the centering effect:
\[
(\mathbf{H}\mathbf{D}^{\mathbf{V}}\mathbf{H})_{ij} = \mathbf{D}^{\mathbf{V}}_{ij}- \frac{1}{n} \sum_{t=1}^{n} \mathbf{D}^{\mathbf{V}}_{it} - \frac{1}{n} \sum_{t=1}^{n} \mathbf{D}^{\mathbf{V}}_{tj}+\frac{1}{n^2} \sum_{s,t=1}^{n} \mathbf{D}^{\mathbf{V}}_{st},
\]
where the whole matrix mean can be computed as 
\[
\frac{1}{n^2} \sum_{i,j=1}^{n} \mathbf{D}^{\mathbf{V}}_{ij}=\beta(1-\frac{\sum_{t=1}^{K}n_{t}^2}{n^2}),
\]
and the mean of each row can be computed as 
\[
\frac{1}{n} \sum_{t=1}^{n} \mathbf{D}^{\mathbf{V}}_{it} = \beta(1-\frac{n_s}{n})
\]
assuming the $i$th point belongs to group $s$. It follows that
\[
(\mathbf{H}\mathbf{D}^{\mathbf{V}}\mathbf{H})_{ij}=\begin{cases}
\beta\left(\frac{2n n_s-\sum_{t=1}^{K} n_{t}^2}{n^2}-1\right)       & \text{if $s=t$},\\
\beta\left(\frac{n (n_s+n_t)-\sum_{t=1}^{K} n_{t}^2}{n^2}\right)  & \text{if $s \neq t$},
\end{cases}
\]
for each group $s$ and $t$. Next, we would like to establish the following equality for the within group entries:
\begin{align}
\label{eq:aux1}
\frac{2n n_s-\sum_{t=1}^{K} n_{t}^2}{n^2}-1 = - \sum_{s \neq t}^{1,\cdots, K}\frac{n (n_s+n_t)-\sum_{l=1}^{K} n_{l}^2}{n^2} \cdot \frac{n_t}{n_s}
\end{align} 
for each group $s$. Without loss of generality, assume $s=1$ and multiply $n^2$ to it. Thus, proving Equation~\ref{eq:aux1} is equivalent to prove
\begin{align*}
&2n n_1-\sum_{t=1}^{K} n_{t}^2-n^2 + \sum_{l=2}^{K}n (n_1+n_t) \frac{n_t}{n_1} -\sum_{t=1}^{K} n_{t}^2 \sum_{l=2}^{K}\frac{n_t}{n_1}=0 \\
\Leftrightarrow & \ 2n n_1 - n^2 +  \sum_{l=2}^{K}n n_t - \sum_{t=1}^{K} n_{t}^2 - \sum_{t=1}^{K} n_{t}^2 \sum_{l=2}^{K}\frac{n_t}{n_1} + \sum_{l=2}^{K}n \frac{n_t^2}{n_1}=0 \\
\Leftrightarrow & \ 2n n_1 - n^2 +  n (n-n_1) - \sum_{t=1}^{K} n_{t}^2 - \sum_{t=1}^{K} n_{t}^2 \frac{n-n_1}{n_1} + \frac{n}{n_1}\left(\sum_{t=1}^{K} n_{t}^2-n_1^2\right) =0 \\
\Leftrightarrow & \ n n_1 - \sum_{t=1}^{K} n_{t}^2\left(1+\frac{n-n_1}{n_1}-\frac{n}{n_1}\right)-n n_1 =0.
\end{align*} 
All terms cancel out at the last step, ensuring the equality of Equation~\ref{eq:aux1}.

Therefore, Equation~\ref{eq:aux1} guarantees the weight in each term of $\Energy_{n_s+n_t}(\mathbf{U}_{s}, \mathbf{U}_{t})$ matches the corresponding weight in $\mathbf{H}\mathbf{D}^{\mathbf{V}}\mathbf{H}$, and it follows that 
\begin{align*}
\Dcov_{n}(\mathbf{U}, \mathbf{V})=& \ \frac{1}{n^2}\sum_{i,j=1}^{n} \mathbf{D}^{\mathbf{U}}_{ij} \cdot (\mathbf{H}\mathbf{D}^{\mathbf{V}}\mathbf{H})_{ij}\\
=& \ \frac{\beta}{n^2}\sum_{1 \leq s<t \leq k} \frac{n (n_s+n_t)-\sum_{l=1}^{K} n_{l}^2}{n^2}\biggl\{2\sum_{i=1}^{n_s}\sum\limits_{j=1}^{n_t}d(u_{i}^{s},u_{j}^{t})\\
& - \frac{n_t}{n_s}\sum\limits_{i,j=1}^{n_s}d(u_{i}^{s},u_{j}^{s})- \frac{n_s}{n_t}\sum\limits_{i,j=1}^{n_t}d(u_{i}^{t},u_{j}^{t})\biggr\} \\
= &\ \beta \sum\limits_{1 \leq s<t \leq k} \frac{n (n_s+n_t)-\sum_{l=1}^{K} n_{l}^2}{n^4} n_s n_t \cdot  \Energy_{n_s,n_t}(\mathbf{U}_{s}, \mathbf{U}_{t}).
\end{align*} 

Finally, comparing the distance covariance:
\[
\begin{aligned}
\Dcov_{n}(\mathbf{U}, \mathbf{V}) = \sum\limits_{1 \leq s<t \leq k}\beta&\left\{ \frac{n (n_s+n_t)-\sum_{l=1}^{K} n_{l}^2}{n^4} n_s n_t  \cdot  \Energy_{n_s,n_t}(\mathbf{U}_{s}, \mathbf{U}_{t}) \right\}
\end{aligned}
\]
to \Disco:
\[
\Disco(\{\mathbf{U}_{k}\}) = \sum\limits_{1 \leq s<t \leq K}\left\{ \frac{n_s n_t}{2n}\Energy_{n_s,n_t}(\mathbf{U}_{s}, \mathbf{U}_{l}) \right\},
\]

We observe that they are simply different weightings of the same pairwise energy statistics, and they are the same up to scaling if and only if $n (n_s+n_t)-\sum_{l=1}^{K} n_{l}^2$ is a fixed constant for all possible $s \neq t$, or equivalently, $n_s+n_t$ is fixed. This is true when either $k=2$ (the trivial case), or $n_1=n_2=\ldots=n_k = \frac{n}{k}$, in which case
\begin{align*}
\Dcov_{n}(\mathbf{U}, \mathbf{V})
&= \frac{\beta}{n^2 k}\sum\limits_{1 \leq s<t \leq k}\{ n^2 \cdot \Energy_{n_s,n_t}(\mathbf{U}_{s}, \mathbf{U}_{t}) \} \\
&=\frac{2\beta}{n k} \Disco(\{\mathbf{U}_{k}\}).
\end{align*} 
This confirms that distance covariance (\Dcov) and the distance components (\Disco) are equivalent up to a scaling factor when the sample sizes $n_k$ are equal across all groups, which can be achieved via bootstrap sampling on given sample data.
\end{proof}

\section{Additional Simulations}
\label{appen:function}

In this section, we consider $20$ additional multivariate simulations, including polynomial (linear, quadratic, cubic), trigonometric (sinusoidal, circular, ellipsoidal, spiral), geometric (square, diamond, W-shaped), and other relationships, based on the same settings as in \cite{MGC}. In each case, we draw $n$ samples from a two-dimensional distribution $F_{U_1}$. Then, $F_{U_2}$ is generated via $F_{U_1}$ but rotated 60 degrees counterclockwise, and $F_{U_3}$ is rotated 60 degrees clockwise relative to $F_{U_1}$. We then test the 3-sample hypothesis that $F_{U_1} = F_{U_2} = F_{U_3}$. 

For the test statistics, we used \Dcov, \Hsic, \Mgc, and  random-forest induced measure (\Kmerf), which are dependence measures, as well as two linear correlations, canonical correlation analysis (\CCA) and RV coefficient (\RV), and the classical \Manova~test. 

\begin{figure*}[htbp]
    \centering
    \includegraphics[width=0.99\linewidth]{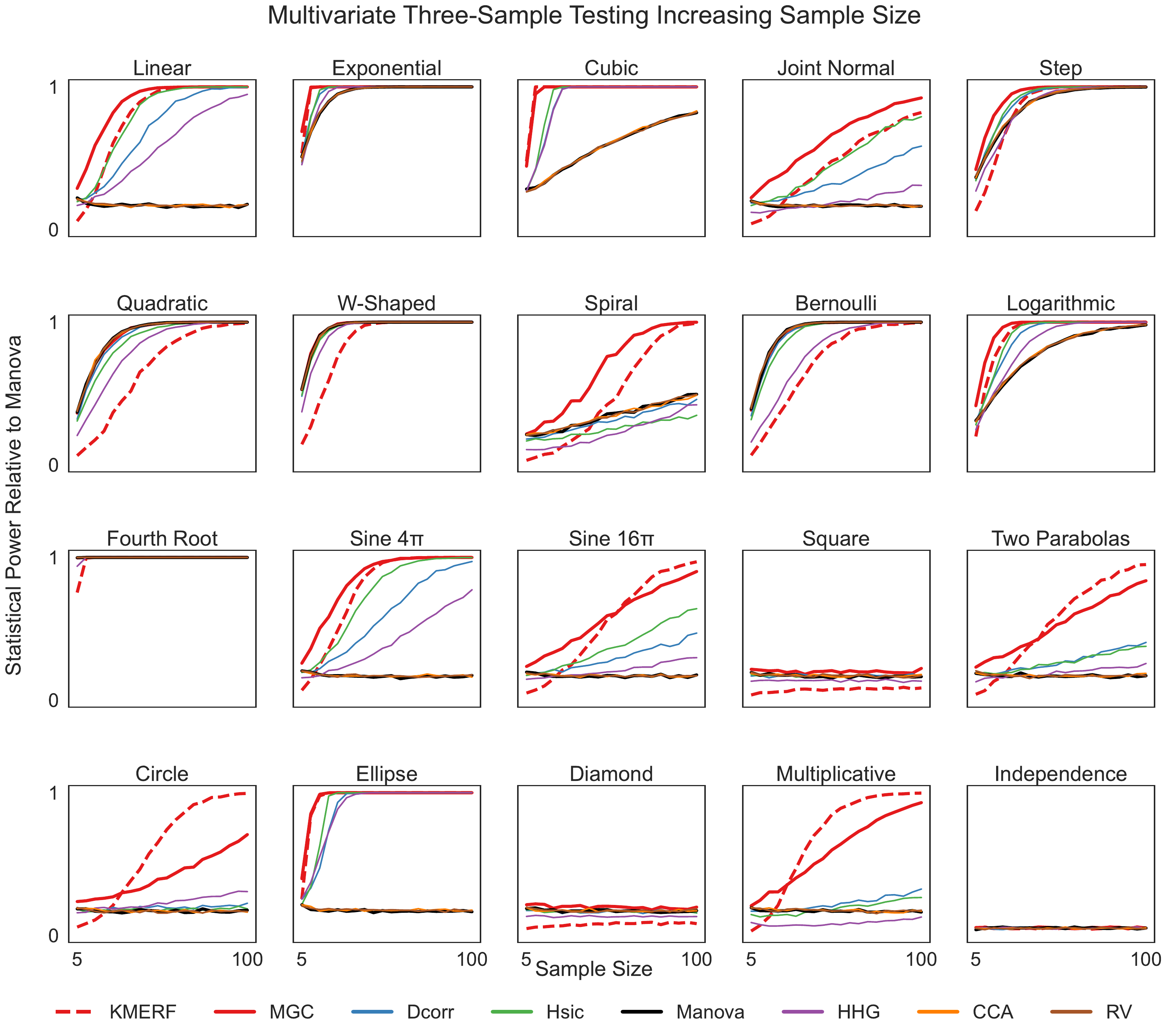}
    \caption{This figure presents the K-sample testing power at a type-1 error level of $0.05$, using several universally consistent dependence measures, two linear correlations, and \Manova, as the sample size increases, for 20 different distribution settings..}
    \label{fig1}
\end{figure*}

Figure~\ref{fig1} shows the testing power of each method in each case, as sample size increases. Since different dependence measures are built on different kernels, metrics, or formulations, they exhibit different finite-sample testing power. However, all four dependence measures show their testing power converging to 1 as the sample size increases in every panel (except for square, diamond, and independence). This indicates that they successfully detect the distributional difference in every case, matching the universal consistency property. In comparison, the standard \Manova~test, as well as the two linear correlations, \CCA~and \RV, do not have their power increasing to 1 in many settings, such as the cases of linear and joint normal. This is expected, as \Manova~only detects mean differences, while the distributions have the same mean in those cases.

Note that the last panel represents independence, which is rotationally invariant such that $F_{U_1} = F_{U_2} = F_{U_3}$. This serves to verify the validity of all tests, and all dependence measures have the testing power equal to the designated type 1 error level $\alpha=0.05$, verifying the validity of using dependence measures for K-sample tests. Moreover, all tests have low power in the square and diamond cases because these distributions are very close to rotation-invariant, though not exactly, making these cases very similar to independence and thus difficult to detect at small sample sizes. 

\section{Function Details}

In this section, we details the simulation functions used in Figure~\ref{fig1}. In each case, we generate two-dimensional sample data $\mathbf{U}_1$ based on the given distribution $F_{U_1}$. Then, $F_{U_2}$ is generated via $F_{U_1}$ but rotated 60 degrees counterclockwise, and $F_{U_3}$ is rotated 60 degrees clockwise relative to $F_{U_1}$. 

In the following, $U_1(1)$ and $U_1(2)$ denote the first and second dimension of the random variable $U_1$, and $\epsilon$ is a white noise from independent standard normal distribution unless mentioned otherwise.

\begin{itemize}
\item Linear:
\begin{align*}
U_1(1) &\sim \mbox{Uniform}(-1,1),\\
U_1(2) &=U_1(1)+\epsilon.
\end{align*}
\item Exponential:
\begin{align*}
U_1(1) &\sim \mbox{Uniform}(0,3), \\
U_1(2) &=\mbox{exp}( U_1(1))+10\epsilon.
\end{align*}
\item Cubic:
\begin{align*}
U_1(1) &\sim \mbox{Uniform}(-1,1), \\
U_1(2) &=128( U_1(1)-\tfrac{1}{3})^3+48( U_1(1)-\tfrac{1}{3})^2-12( U_1(1)-\tfrac{1}{3})+80\epsilon.
\end{align*}
\item Joint normal: Let $\Sigma = \begin{bmatrix} 1&0.5\\ 0.5& 1.5 \end{bmatrix}$, then
\begin{align*}
(U_1(1),U_1(2)) &\sim \mbox{Normal}(0, \Sigma).
\end{align*}
\item Step Function:
\begin{align*}
U_1(1) &\sim \mbox{Uniform}(-1,1),\\
U_1(2) &=1( U_1(1)>0)+\epsilon.
\end{align*}
\item Quadratic:
\begin{align*}
U_1(1) &\sim \mbox{Uniform}(-1,1),\\
U_1(2) &=( U_1(1))^2+0.5\epsilon.
\end{align*}
\item W Shape:  $W \sim \mbox{Uniform}(-1,1)$,
\begin{align*}
U_1(1) &\sim \mbox{Uniform}(-1,1),\\
U_1(2) &=4\left[ \left( ( U_1(1))^2 - \tfrac{1}{2} \right)^2 +  W/500 \right]+0.5\epsilon.
\end{align*}
\item Spiral: $W \sim \mbox{Uniform}(0,5)$,
\begin{align*}
U_1(1)&=W \cos(\pi W),\\
U_1(2)&= W \sin(\pi W) +0.4 \epsilon.
\end{align*}
\item Uncorrelated Bernoulli: $W \sim \mbox{Bernoulli}(0.5)$, $\epsilon_{1} \sim \mbox{Normal}(0, 1)$,
\begin{align*}
U_1(1) &\sim \mbox{Bernoulli}(0.5)+0.5\epsilon,\\
U_1(2)&=(2W-1) U_1(1)+0.5\epsilon_{1}.
\end{align*}
\item Logarithmic:
\begin{align*}
U_1(1) &\sim \mbox{Normal}(0, 1),\\
U_1(2)&=2\log_{2}(|U_1(1)|)+3\epsilon.
\end{align*}
\item Fourth Root:
\begin{align*}
U_1(1) &\sim \mbox{Uniform}(-1,1),\\
U_1(2)&=| U_1(1)|^\frac{1}{4}+\frac{1}{4}\epsilon.
\end{align*}
\item Sine Period $4\pi$: $\theta=4\pi$,
\begin{align*}
U_1(1)&\sim \mbox{Uniform}(-1,1), \\
U_1(2) &=\sin ( \theta U_1(1) )+0.02 \epsilon.
\end{align*}
\item Sine Period $16\pi$: Same as above except $\theta=16\pi$ and the noise on $U_1(2)$ is changed to $0.5\epsilon$.
\item Square: Let $W_1 \sim \mbox{Uniform}(-1,1)$, $W_2 \sim \mbox{Uniform}(-1,1)$, $\theta=-\frac{\pi}{8}$. Then
\begin{align*}
U_1(1)&=W_1 \cos\theta + W_2 \sin\theta + 0.05 \epsilon,\\
U_1(2)&=-W_1 \sin\theta + W_2 \cos\theta.
\end{align*}
\item Two Parabolas: $W \sim \mbox{Bernoulli}(0.5)$,
\begin{align*}
U_1(1) &\sim \mbox{Uniform}(-1,1),\\
U_1(2) &=\left( ( U_1(1))^2  + 2\epsilon\right) \cdot (W-\tfrac{1}{2}).
\end{align*}
\item Circle: $W \sim \mbox{Uniform}(-1,1)$, $r=1$,
\begin{align*}
U_1(1)&=r \left(\cos(\pi W)+0.4 \epsilon \right),\\
U_1(2)&= \sin(\pi W).
\end{align*}
\item Ellipse: Same as above except $r=5$.
\item Diamond: Same as  ``Square'' except $\theta=-\frac{\pi}{4}$.
\item Multiplicative Noise: 
\begin{align*}
U_1(1) &\sim \mbox{Normal}(0, 1),\\
U_1(2) &= \epsilon U_1(1).
\end{align*}
\item Independence: Let $W_1 \sim \mbox{Normal}(0,1)$, $W_2 \sim \mbox{Normal}(0,1)$, $W_3 \sim \mbox{Bernoulli}(0.5)$, $W_4 \sim \mbox{Bernoulli}(0.5)$. Then
\begin{align*}
U_1(1)&=W_1/3+2W_3-1,\\
U_1(2)&=W_2/3+2W_4-1.
\end{align*}
\end{itemize}

\begin{figure}
    \begin{center}
    \includegraphics[width=0.8\linewidth]{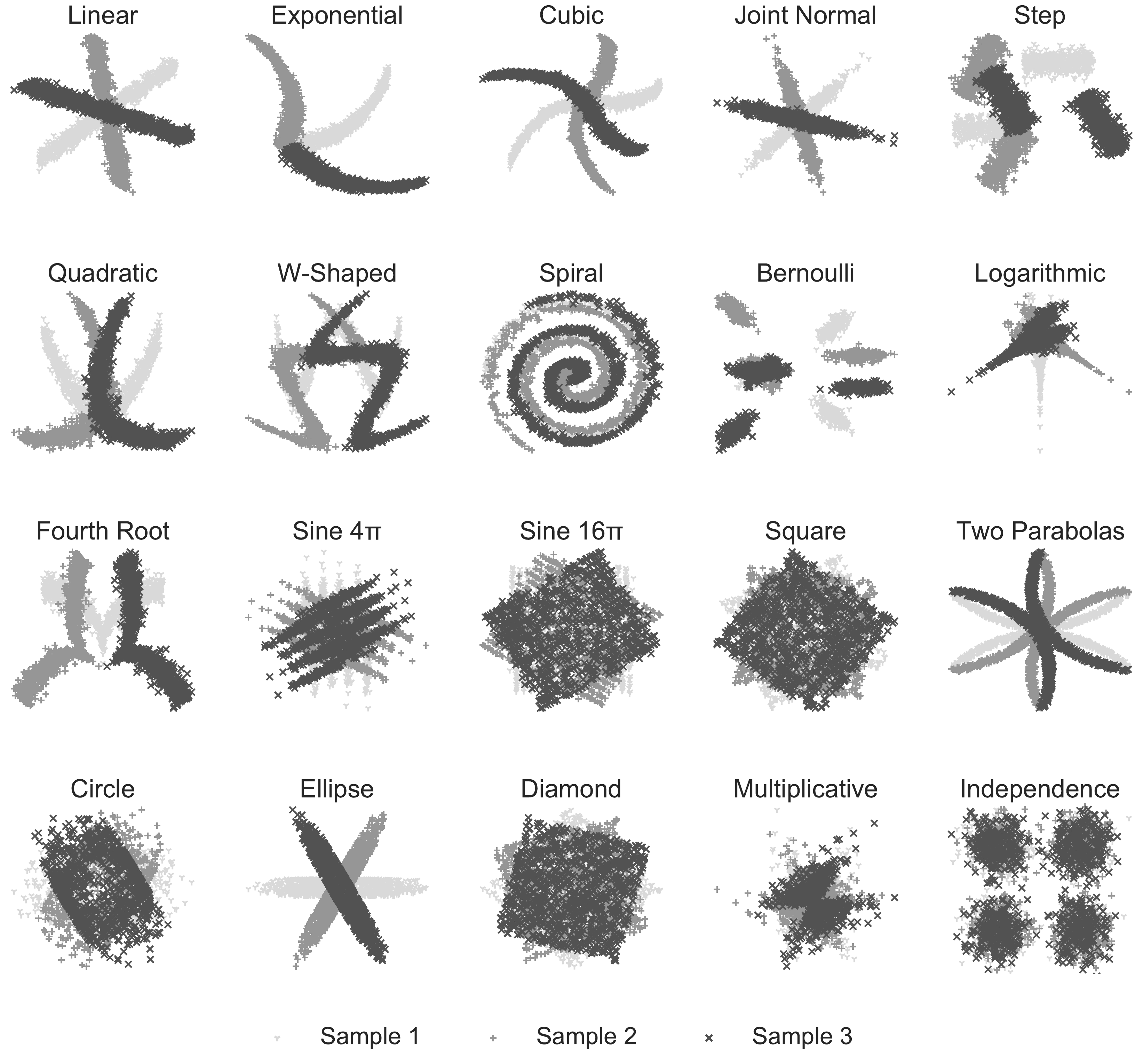}
    \end{center}
    \caption{This figure visualizes the 2-dimensional distributions used in Figure~\ref{fig1}. We sample $500$ points from the first distribution $F_{U_1}$ (black dots), then rotate 60 degrees clockwise to produce $F_{U_2}$ and 60 degrees counter-clockwise to produce $F_{U_3}$, marked by lighter dots in each case.}
    \label{fig3}
\end{figure}

The sample data visualizations are then provided in Figure~\ref{fig3}. It is clear that the 3-sample tests for each panel should be significant, except in the independence case where rotations do not change the distribution, and in the square/diamond cases, which are very difficult as most of the points appear to stay similar to the original distribution. Moreover, cases like linear, joint normal, and ellipse have different distributions but the same mean, such that \Manova~is expected to underperform and be unable to detect the distributional differences.

\end{document}